# Use of natural language processing to extract and classify papillary thyroid cancer features from surgical pathology reports

**Authors:**

Ricardo Loor-Torres, MD*[1]; Yuqi Wu, PhD*[3]; Esteban Cabezas, MD[1]; Mariana Borras, MD[1]; David Toro-Tobon, MD[2]; Mayra Duran, MD[1]; Misk Al Zahidy, MS[1]; Maria Mateo Chavez, MD[1]; Cristian Soto Jacome, MD[1]; Jungwei W. Fan, PhD[3]; Naykky M. Singh Ospina, MD[4]; Yonghui Wu, PhD[5]; Juan P. Brito, MD[1,2]

**Author affiliations:**

1. Knowledge and Evaluation Research Unit, Division of Endocrinology, Diabetes, Metabolism, and Nutrition, Department of Medicine, Mayo Clinic, Rochester, Minnesota
2. Division of Endocrinology, Diabetes, Metabolism, and Nutrition, Mayo Clinic, Rochester, Minnesota
3. Department of Artificial Intelligence and Informatics, Mayo Clinic, Rochester, Minnesota
4. Division of Endocrinology, Department of Medicine, University of Florida, Gainesville, Florida
5. Department of Health Outcomes and Biomedical Informatics, University of Florida, Gainesville, Florida

* Shared first authorship

**Corresponding author:**

Juan P. Brito MD
Division of Endocrinology, Diabetes, Metabolism, and Nutrition
Mayo Clinic, Rochester, Minnesota
200 First Street SW
Rochester, MN 55902.
(507) 266-0500
brito.juan@mayo.edu

## Abstract

### Background

We aim to use Natural Language Processing (NLP) to automate the extraction and classification of thyroid cancer risk factors from pathology reports.

### Methods

We analyzed 1,410 surgical pathology reports from adult papillary thyroid cancer patients at Mayo Clinic, Rochester, MN, from 2010 to 2019. Structured and non-structured reports were used to create a consensus-based ground truth dictionary and categorized them into modified recurrence risk levels. Non-structured reports were narrative, while structured reports followed standardized formats. We then developed ThyroPath, a rule-based NLP pipeline, to extract and classify thyroid cancer features into risk categories. Training involved 225 reports (150 structured, 75 unstructured), with testing on 170 reports (120 structured, 50 unstructured) for evaluation. The pipeline's performance was assessed using both strict and lenient criteria for accuracy, precision, recall, and F1-score.

### Results

In extraction tasks, ThyroPath achieved overall strict F-1 scores of 93% for structured reports and 90 for unstructured reports, covering 18 thyroid cancer pathology features. In classification tasks, ThyroPath-extracted information demonstrated an overall accuracy of 93% in categorizing reports based on their corresponding guideline-based risk of recurrence: 76.9% for high-risk, 86.8% for intermediate risk, and 100% for both low and very low-risk cases. However, ThyroPath achieved 100% accuracy across all thyroid cancer risk categories with human-extracted pathology information.

### Conclusions

ThyroPath shows promise in automating the extraction and risk recurrence classification of thyroid pathology reports at large scale. It offers a solution to laborious manual reviews and advancing virtual registries. However, it requires further validation before implementation.

## Introduction

Thyroid cancer has become the most prevalent cancer among American adults aged 16-33 years, ranking as the sixth most common cancer in women and the second most common cancer in Hispanic women.[1,2] Despite its increasing incidence, the mortality rate for thyroid cancer remains low and stable, with approximately 0.5 deaths per 100,000 individuals.[3] Therefore, the focus of thyroid cancer research and care, particularly for well-differentiated types, such as papillary, follicular, and oncocytic thyroid cancer, has shifted from predicting survival to identifying recurrence risk.[4,5]

The accurate risk stratification of thyroid cancer recurrence requires critical information from pathology reports, such as tumor histology and tumor extension.[6] The curation of this data is primarily a manual task, with clinicians evaluating their patients' recurrence risk based on established clinical guideline frameworks. One of the most well-known frameworks is the risk classification system proposed by the American Thyroid Association (ATA) in 2015.[7] This system enables clinicians to categorize patients as low, intermediate, or high risk, facilitating for personalized treatment and follow-up strategies.

However, the use of pathology features and subsequent risk classification has been limited at the population level, with only a few institutions manually creating registries for research.[8] These tasks could be both financially burdensome and time-consuming, a barrier that might hinder the use of thyroid pathology variables in multicenter and national registries. There is a pressing need to address this bottleneck to facilitate studies leveraging extensive thyroid pathology data, which could significantly impact patient care. This involves evaluating pathology reporting quality, deepening understanding of thyroid cancer epidemiology and management, and refining the classification of recurrence risk.[9,10] These efforts have the potential to lay the groundwork for designing interventions and evaluating their efficacy through both retrospective and prospective studies across various centers.

Artificial Intelligence (AI) has emerged as a valuable tool for analyzing large and complex datasets effectively.[11,12] Natural Language Processing (NLP), as the intersection of AI and linguistics, possesses the ability to understand both written and spoken human language.[13] NLP methods have demonstrated their potential in efficiently extracting and standardizing data for easy utilization in clinical and research settings.[14,15] These methods have been evaluated and implemented in various fields of medicine, such as Radiology and Oncology, particularly in tasks focused on extracting and categorizing information from electronic health records (EHRs).[16,17]

In thyroidology, the exploration of NLP's potential applications, particularly in thyroid cancer, has been increasing in recent years.[12,18,19] A previous study explored the use of NLP to extract key elements from thyroid cancer pathology reports.[10] However, these applications have not addressed the risk of recurrence. Recognizing the need for novel approaches to harness thyroid cancer data, our goal was to develop an NLP pipeline, "ThyroPath", that extract features, and subsequently classify guideline-based recurrence risk from surgical pathology reports.

## Methods

**Ethical considerations:**

This study received approval from the Mayo Clinic Institutional Review Board. Only patients who had granted Minnesota (MN) research authorization were included in our study.

**Data source**

We included de-identified surgical pathology reports from adult patients (aged >18 years) who underwent primary thyroid surgery at Mayo Clinic in Rochester, MN, from January 1, 2010, to December 31, 2019. As an initial step in the development of the NLP pipeline, we focused only on patients with available surgical pathology results who were diagnosed with papillary thyroid cancer (PTC), as this cancer type represents almost 90% of all thyroid malignancies.[20] We also included structured and non-structured pathology reports. Non-structured cases relied more on free text and narrative formats, exhibiting variability in reporting variables. For instance, core elements were sometimes missing from the report, potentially relying on the pathologist's criteria. Structured reports followed the College of American Pathologists (CAP) recommendations for thyroid pathology synoptic reporting, and these reports have standardized forms for pathology feature elements. Mayo Clinic implemented CAP reporting guidelines in 2017.

Initially, 1,410 cases were queried and categorized into structured (n=270) and unstructured (n=1,140) cohorts. All available structured cases were used, while 125 unstructured cases were randomly selected. The training set for the NLP classification pipeline development comprised 225 reports, including 150 structured and 75 unstructured. The evaluation was conducted on the remaining 170 independent test reports, including 120 structured and 50 unstructured.

**Ground-truth development**

Guided by thyroid specialists, our team developed an annotation dictionary **(Supplemental Table 1)** based on CAP recommendations, highlighting core and conditional (reported according to institutional protocol) elements. Using the biomedical annotation system MedTator **(Figure 1)**, we annotated categories (e.g., tumor site) with respective attributes (e.g., left thyroid lobe) to establish a consensus-based ground truth. Following the 2015 ATA guidelines for differentiated thyroid cancer recurrence risk stratification,[7] we developed an annotation flowchart **(Supplemental Figure 1)** with four predefined guideline-based categories: Very low, low, indeterminate, and high risk. Notably, we introduced a "very low risk" category for well-differentiated thyroid microcarcinomas (<10 mm) that excluded any additional features such as aggressive histologies or lymph node metastasis. Three reviewers (MB, EC, RL) underwent a calibration process, evaluating 20-30 reports per session. They achieved a pre-established kappa agreement rate of over 80% at each project stage, ensuring consistency and precision.

**ThyroPath pipeline overview**

ThyroPath is a multi-step, rule-based NLP pipeline created specifically for extraction of pathology features and subsequent guideline-based categorization of thyroid cancer recurrence risk from surgical pathology reports. The initial step involved the extraction task, followed by the classification task as the subsequent step. The extraction process involved tasks such as identifying categories and retrieving linked attributes **(Figure 2)**. Additionally, the extracted variables underwent further processing based on a pre-defined risk-labeled categorization **(Figure 3)**, leading to the classification of each case under the four categories: very low, low, intermediate, and high risk.

The NLP extraction task was developed and tested in both structured and unstructured pathology reports. However, the NLP classification tasks was developed and tested only in structured reports. The consideration of employing unstructured reports for the classification task was also explored. However, they were ultimately disregarded due to significant heterogeneity and a lack of standardized reporting on specific features. For instance, some reports provided a final interpretation without detailing crucial aspects like the size of metastatic foci or extrathyroidal extension. This resulted in insufficient data for certain features and hindered our ability to estimate the guideline-based risk of cancer recurrence.

**Development of NLP extraction task**

To analyze the data, our NLP strategy employed two methods to segment the report based on specific criteria, enabling the retrieval of more granular information. For structured reports, we applied header-based topic segmentation **(Supplemental Figure 2)**, where a set of keywords was used to identify categories (e.g., tumor focality) and their linked attributes (e.g., unifocal) for retrieval and extraction. This approach relied on the principle that each sentence within structured reports corresponds to a specific category and its respective attributes.

Conversely, unstructured reports presented a distinct challenge due to their lack of consistency in reporting and their more free-text nature. To address this issue, we employed a topic segmentation approach **(Supplemental Figure 2)**, which identified keywords throughout the entire text to anchor and extract information. Additionally, in cases where specific keywords (e.g., tumor size) were absent, we created a set of key strings (e.g., "tumor measuring" or "greatest dimension of") as surrogates to locate the relevant data, specifically for numerical variables **(Supplemental Table 2)**.

**Development of NLP guideline-based risk classification task**

We employed a hierarchical approach where our classifier prioritized detecting higher-risk features. For instance, if any high-risk feature was identified, cases were automatically classified as such, regardless of other features. Conversely, in the absence of high-risk features but with intermediate risk features present, reports were categorized as intermediate risk. Subsequently, reports lacking both high and intermediate risk indicators underwent evaluation for low-risk features. Ultimately, reports lacking risk indicators at any level were classified as very low risk. This systematic approach ensured a comprehensive assessment of risk levels for each case.

**Evaluation of the NLP extraction task**

We compared the NLP pipeline's extracted features against the ground truth dictionary developed by human annotation. The performance of the pipeline was evaluated using both strict (e.g., rigid, or exact word matching) and lenient (e.g., flexible, or partial word matching) criteria for accuracy, precision, recall, and F1-score. Accuracy is defined as the correct identification of attributes on the pathology report by the machine in all instances out of the total predictions made. Precision represents the proportion of positive identifications that were accurate, while recall indicates the proportion of actual positives correctly identified. The F-1 score is a standard performance metric commonly used to assess extraction tasks within NLP methodologies, providing a balanced assessment of model precision and recall.

**Evaluation of the guideline-based risk classification task**

We conducted a comparison between the classifications generated by the NLP pipeline and those derived from human-annotated ground truth reached through consensus. To facilitate this comparison, we utilized a 2 x 2 confusion matrix. We calculated the overall accuracy by determining the ratio of true positives across all categories to the total number of cases. Moreover, we assessed the accuracy within each specific risk category. True positives were defined as cases where the machine classification aligned with the ground truth for that particular category. The accuracy for each category was then computed by dividing the number of true positive predictions by the total number of cases within that category. Furthermore, we examined the frequency of significant discrepancies between the NLP and ground truth. For example, instances where the NLP classified a case as high risk when it was actually very low or low risk, or vice versa.

## Results

### Cohort

A total of 395 surgical pathology reports were included in the present study, comprising 270 structured and 125 unstructured reports. Of these, 55.96% (150 structured and 75 unstructured) were used to develop the NLP extraction pipeline. The remaining 43.04% (120 structured and 50 unstructured) were used for testing the pipeline's performance. Additionally, the NLP classification task was evaluated using the 120 structured pathology reports from the testing set.

### NLP extraction performance

In **Table 1**, we display a comprehensive overview of the performance metrics for our NLP pipeline, including both strict and lenient measures of accuracy, precision, recall, and F-1 scores. The overall strict F1-score was 93% for structured reports and 90% for unstructured reports. Upon single variable analysis, we noted that 11 out of 17 categories from the structured reports achieved an F1-score over 90%, with only the number of lymph nodes involved displaying scores below 80%. Within the unstructured reports, most variables exhibited F1-score of 80% and above, except for margins and extranodal extension, which exhibited scores of 70% and 62%, respectively **(Supplemental table 3)**.

### Guideline-based risk classification performance

The distribution of risk categories by human annotation was as follows: High risk (n=13, 10.83%), intermediate risk (n=38, 31.66%), low risk (n=62, 51.66%), and very low risk (n=7, 5.83%). The guideline-based risk classification pipeline using NLP-extracted features from pathology reports achieved an overall accuracy of 93.3% **(Table 2)**. The accuracy rates for identifying high-risk categories were 76.92%, for intermediate-risk 86.84%, for low risk 100%, and for very low risk 100%. Notably, when using thyroid pathology information extracted by human annotators instead of the ThyroPath, our classifier achieved an overall accuracy of 100% for all the categories **(supplemental table 4)**. This validates the reliability of the classification rules, indicating that any classification errors are likely due to input issues, such as missing or incorrect NLP extraction.

From the error analysis, we noted a significant discrepancy in a case classified as high-risk by the ground truth but as low risk by the rule-based pipeline **(Figure 4).** In addition, our pipeline successfully identified entities specific to each use case category classification. For instance, extranodal extension was the primary factor for classifying cases as high risk, with a prevalence of 100% (n=10). In the intermediate risk category, the primary factors were the number of lymph nodes > 5 (n=18, 51.42%), followed by tumor size > 4 cm (n=11, 31.42%), angioinvasion (n=5, 14.28%), and histologic subtypes (n=1, 2.85%). Within the low-risk category, the most prevalent attribute was tumor size between 4 and 1 cm (n=54, 79.41%), followed by the presence of fewer than 5 lymph nodes (n=14, 20.58%). The remaining cases were classified as very low risk, with no annotated variables and a tumor size of less than 1 cm. The distribution of the complete human annotated cohort of structures reports (n=270), categorized according to their risk level entities, is displayed in **Supplemental Table 5**.

**Discussion**

In this study, we developed a rule-based NLP pipeline, ThyroPath, to extract crucial attributes from papillary thyroid cancer pathology reports and accurately assign each use case according to the appropriate guideline-based risk category.

Presently, and to the best of our knowledge, two studies have reported the use of NLP tools in thyroid cancer pathology reports. Lee et al. published a preprint report describing the development of a local large language model (LLM) using a question-and-answer approach. The model identified and extracted key data from surgical pathology reports for thyroid cancer staging and risk of recurrence assessment. Their findings demonstrated that the model performed comparably to humans and highlighted the expected reduction in time required for this assessment.[21] Furthermore, Yoo et al. successfully extracted AJCC/TNM staging data from a substantial dataset of 63,795 thyroid cancer pathology reports and standardized it into a common data dictionary.[10] However, our study is the first to develop an NLP pipeline, that extracts 18 clinically and research-relevant pathology variables from thyroid cancer pathology variables from thyroid cancer reports with high granularity and specificity. Moreover, we have successfully used the NLP-extracted information to assign guideline-based recurrence risk classifications to each report. Our approach offers superb interpretability, allowing users to trace the attributes used for classification.

During the development of ThyroPath, we encountered significant challenges, particularly regarding the evolving structure of pathology reports. We found that ThyroPath's performance was affected by the level of structure present in each pathology report. Unstructured reports, with their varied language and extent of parameters reported, presented substantial obstacles to entity recognition and relation extraction.[22] While no technique can retrieve non-reported variables from the original text, advancements in AI, particularly LLMs, can understand the context of reports and might improve the accuracy of extracting information from variably structured or unstructured reports.[23] Additionally, a considerable proportion of granular data was missing which also precluded us to its use for risk of recurrence classification purposes. This observation holds implications for clinical documentation practices, as most pathology reports used to lack structure prior to the adoption of reporting guidelines. Additionally, a considerable proportion of granular data was missing within the unstructured reports, which precluded its use for risk of recurrence classification purposes. This observation has implications for clinical documentation practices, as most pathology reports lacked structure prior to the adoption of reporting guidelines. Due to this limitation, we restricted our classification tasks to the dataset of structured reports that included all necessary elements for guideline-based risk classification.

By leveraging well-known rule-based NLP techniques such as entity recognition and classification tasks, ThyroPath produces highly interpretable outputs. For example, we can discern which attribute drove the guideline-based classification, a feature crucial for interpretability in medical NLP applications. Moreover, the use of a rule-based approach enables us to adapt classification rules for future changes from the guidelines, enhancing ThyroPath's flexibility.

Despite the strengths of our study and methods, several limitations should be considered. Our NLP component is grounded in data from a single institution, potentially limiting its generalizability to pathology reports from different institutions with varying vocabularies or report structures. For instance, our study employed a specific segmentation technique for extracting relevant pathology report information, which may require adjustments in preprocessing steps and re-validation of model performance across diverse settings. Further external validation using diverse data sources and formats is necessary. Subsequently, fine-tuning of our pipeline is essential to enhance its performance and transferability. Additionally, ThyroPath currently focuses on processing information from PTC. This decision aligns with the pilot stage of our project and the high frequency of this cancer, which provides ample data to bootstrap the initial development. However, we are actively working on the next version of ThyroPath to extend its capabilities to extract information from other thyroid cancer subtypes.

Our work has implications for research and practice. The successful development and validation of ThyroPath, our NLP pipeline for extracting pathology features and assigning guideline-based classifications from thyroid cancer reports, paves the way for future research and clinical applications. By automating the extraction of essential data from pathology reports, ThyroPath can streamline the process of screening and data collection for pragmatic studies and clinical trials, ultimately accelerating research efforts in the field of thyroid cancer. Moreover, the tool has the potential to contribute to the establishment of extensive digital thyroid cancer registries powered by AI. This can aid in comprehending more intricate disease trends and in the development of improved predictive models or treatment strategies for thyroid cancer recurrence. Furthermore, a validated ThyroPath could prove invaluable for clinical practice, especially considering that most clinicians in the United States are not thyroid experts. Many healthcare professionals may lack extensive familiarity with thyroid cancer guideline practices. ThyroPath's provision of clear risk classification categories based on pathology reports can empower non-specialist clinicians to make informed decisions regarding patient management, bridging the gap between clinical practice and the complexities of pathology data. ThyroPath may promote standardized and evidence-based clinical decisions, enhancing patient care and reducing variability in risk assessment.

**Conclusion**

Our study demonstrates the effectiveness of ThyroPath, an NLP-enabled pipeline for extracting pathology features and assigning a guideline-based recurrence risk classification of papillary thyroid cancer reports. This research contributes to the growing body of evidence supporting the utility of NLP and AI techniques in healthcare. However, it's important to note that ThyroPath requires external validation before widespread clinical use. Once validated, it holds the potential to support clinical trials, registry development, and pragmatic trials, thereby advancing thyroid cancer research and improving patient care.

**Disclosures**

None of the authors have any relevant disclosures.

**Funding**

Juan P. Brito and Cristian Soto Jacome were supported by the National Cancer Institute of the National Institutes of Health under Awards Numbers R37CA272473. Naykky Singh Ospina was supported by the National Cancer Institute of the National Institutes of Health under Award Number K08CA248972. Yonghui Wu was supported by Patient-Centered Outcomes Research Institute® (PCORI) under Award Number ME-2018C3-14754 and National Institute on Aging under Award Number R56AG069880. The content is solely the responsibility of the authors and does not necessarily represent the official views of the National Institutes of Health, the Patient-Centered Outcomes Research Institute (PCORI) or National Institute on Aging.

**Acknowledgements**

Figures were designed and generated using BioRender.com.

## Tables

### Table 1. Detailed performance of the NLP pipeline for extraction tasks of the categories in the surgical pathology reports, using strict method

| Categories in Pathology Reports | Structured Pathology Reports | | | | Unstructured Pathology Reports | | | |
|---|---|---|---|---|---|---|---|---|
| | Accuracy | Precision | Recall | F1-score | Accuracy | Precision | Recall | F1-score |
| Procedure | 0.69 | 0.86 | 0.78 | 0.82 | 0.70 | 0.92 | 0.74 | 0.82 |
| Tumor Focality | 0.98 | 1.0 | 0.98 | 0.99 | 0.94 | 0.98 | 0.96 | 0.97 |
| Tumor Site | 0.77 | 0.87 | 0.87 | 0.87 | 0.72 | 0.90 | 0.79 | 0.84 |
| Tumor Size | 1.0 | 1.0 | 1.0 | 1.0 | 1.0 | 1.0 | 1.0 | 1.0 |
| Histologic Subtype | 0.66 | 0.81 | 0.78 | 0.80 | 0.74 | 0.88 | 0.82 | 0.85 |
| Margins | 0.96 | 1.0 | 0.96 | 0.98 | 0.54 | 1.0 | 0.54 | 0.70 |
| Angioinvasion (Vascular Invasion) | 0.96 | 0.99 | 0.97 | 0.98 | 0.96 | 0.98 | 0.98 | 0.98 |
| Lymphatic Invasion | 0.95 | 0.98 | 0.97 | 0.97 | 0.88 | 0.94 | 0.94 | 0.94 |
| Lymphovascular Invasion | (n.a.) * | | | | 0.83 | 0.93 | 0.88 | 0.91 |
| Extrathyroidal Extension | 0.90 | 0.96 | 0.94 | 0.95 | 0.90 | 0.96 | 0.94 | 0.95 |
| Number of Lymph Nodes Involved | 0.61 | 0.77 | 0.74 | 0.75 | 0.68 | 1.0 | 0.68 | 0.81 |
| Number of Lymph Nodes Examined | 0.70 | 0.92 | 0.75 | 0.83 | 0.90 | 1.0 | 0.90 | 0.95 |
| Size of Largest Metastatic Deposit | 0.81 | 0.98 | 0.83 | 0.90 | 0.92 | 0.98 | 0.94 | 0.96 |
| Extranodal Extension | 0.95 | 0.99 | 0.96 | 0.97 | 0.45 | 0.63 | 0.61 | 0.62 |
| Pathologic Staging | 1.0 | 1.0 | 1.0 | 1.0 | 1.0 | 1.0 | 1.0 | 1.0 |
| Primary Tumor TNM | 0.98 | 1.0 | 0.98 | 0.99 | 1.0 | 1.0 | 1.0 | 1.0 |
| Lymph Nodes TNM | 0.98 | 1.0 | 0.98 | 0.99 | 0.96 | 0.98 | 0.98 | 0.98 |
| Distant Metastasis | 0.98 | 1.0 | 0.98 | 0.99 | 0.92 | 0.96 | 0.96 | 0.96 |
| **Overall** | 0.88 | 0.95 | 0.91 | **0.93** | 0.84 | 0.95 | 0.87 | **0.90** |

* Lymphovascular invasion is only present in unstructured reports, thus not applicable (n.a.) in structured cohort.

### Table 2. Classification tasks assessing NLP pipeline-extracted features versus ground truth.

| Ground truth \ NLP pipeline | High risk | Intermediate risk | Low risk | Very low risk | Total |
|---|---|---|---|---|---|
| High risk | **10** | 2 | 1 | 0 | 13 |
| Intermediate risk | 0 | **33** | 5 | 0 | 38 |
| Low risk | 0 | 0 | **62** | 0 | 62 |
| Very low risk | 0 | 0 | 0 | **7** | 7 |
| Total | 10 | 35 | 68 | 7 | **120** |

# Figures

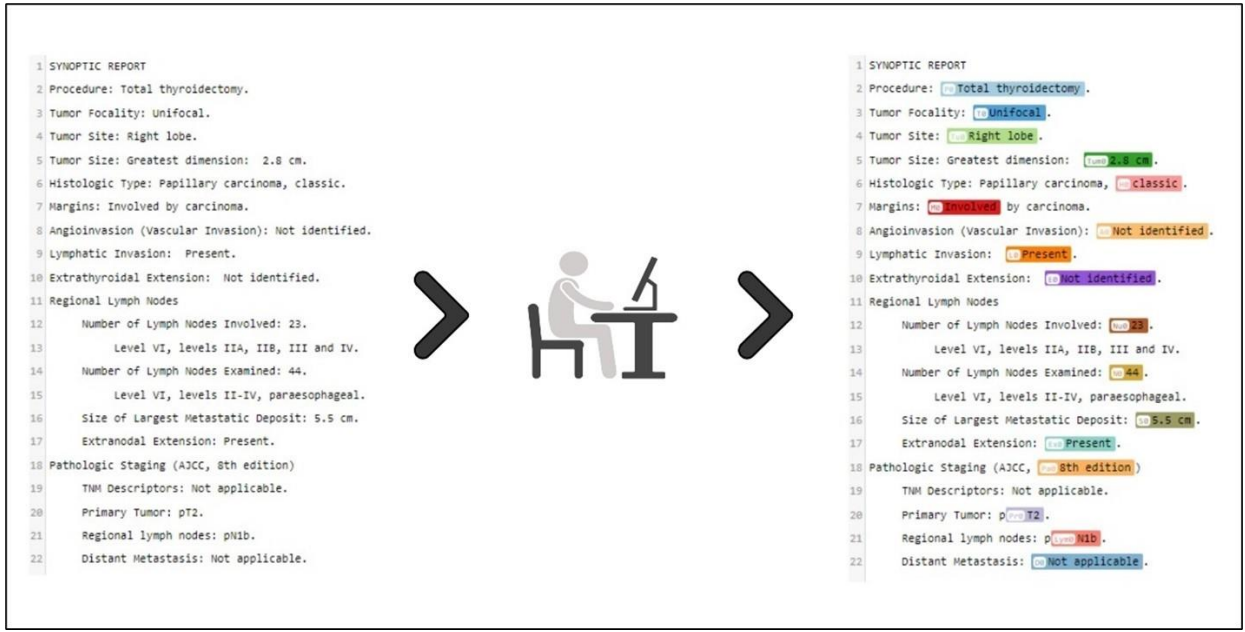

**Figure 1. Annotation of the categories and the corresponding attributes by the MedTator tool.**

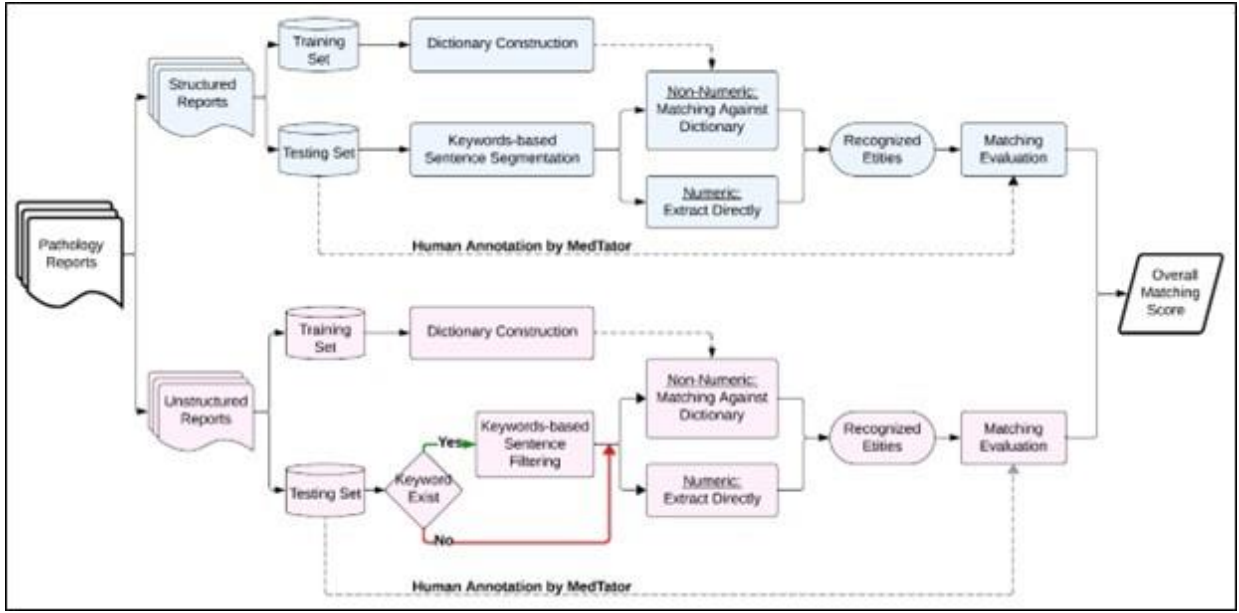

**Figure 2. NLP pipelines for entity recognition strategy for structured and unstructured reports. The process includes steps from initial segmentation of reports, dictionary construction and overall matching score.**

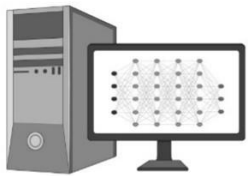

Figure 3. NLP pipeline extraction and classification tasks.

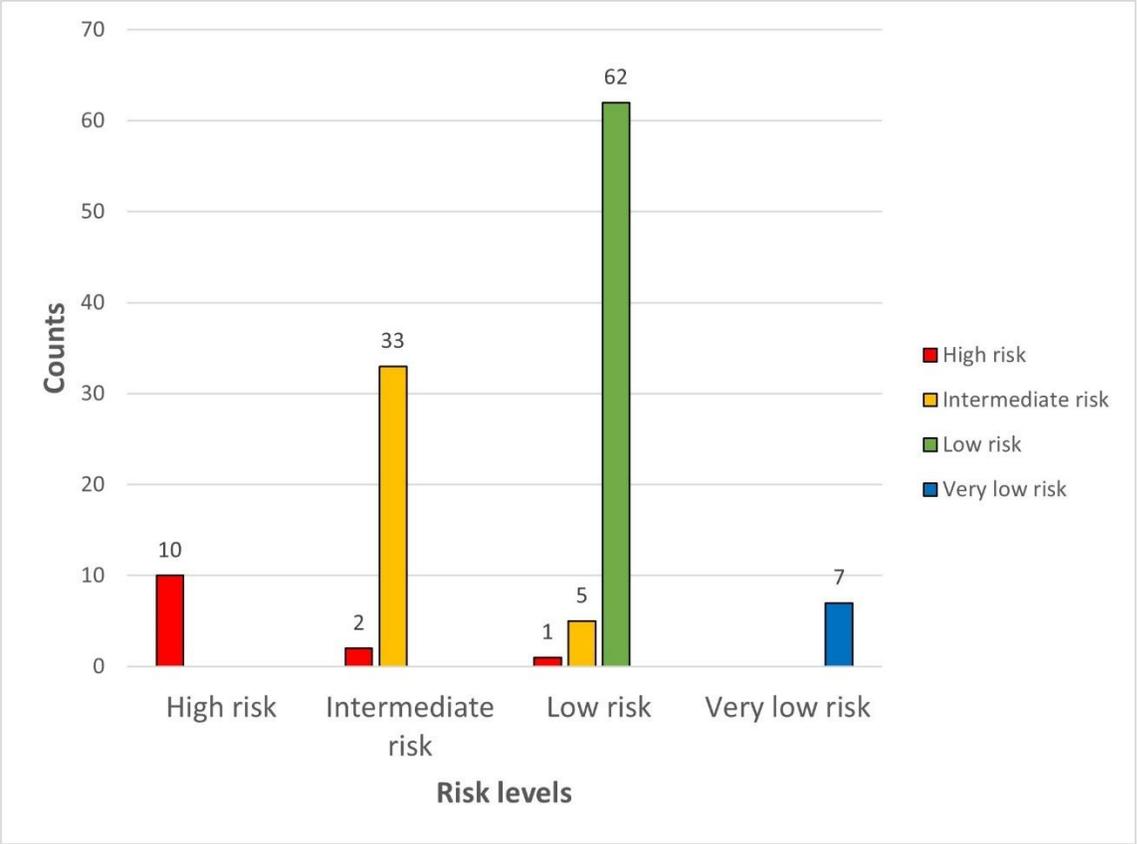

**Figure 4. Frequency and distribution of classification errors comparing NLP-extracted features against Ground Truth.**

*The X-axis represents the ground truth, while the legends represent machine classification.

**Supplemental materials**

| \multicolumn{2}{l}{**Supplemental Table 1. Annotation dictionary with categories and attributes from surgical thyroid pathology reports**} |
|---|---|
| Category | Attributes |
| **Procedure** | Total thyroidectomy (e.g., total thyroidectomy, near-total thyroidectomy)<br>Hemithyroidectomy (e.g., left lobectomy, right lobectomy with or without isthmusectomy)<br>Subtotal thyroidectomy (e.g., right lobectomy with partial left lobectomy)<br>Isthmusectomy |
| **Tumor Focality** | Unifocal<br>Multifocal |
| **Tumor Site** | Right lobe<br>Left lobe.<br>Isthmus |
| **Tumor Size** | Any value in number or letters (e.g., 3 cm or three centimeters) |
| **Histologic Subtype** | PTC, classic variant (e.g., usual, conventional)<br>PTC, follicular variant infiltrative<br>PTC, follicular variant encapsulated<br>PTC, encapsulated variant<br>PTC, microcarcinoma (with or without any other subtype)<br>PTC, infiltrative follicular<br>PTC, tall cell variant<br>PTC, hobnail variant<br>PTC, columnar cell variant<br>PTC, solid / trabecular variant<br>PTC, cribriform-morular variant<br>PTC, diffuse sclerosing variant<br>PTC, warthin-like<br>PTC, oncocytic |
| **Margins** | <u>Positive</u> (e.g., involved by carcinoma, carcinoma present at margin)<br><u>Negative</u> (e.g., negative, uninvolved by carcinoma, free of tumor) |
| **Angioinvasion (Vascular Invasion)** | <u>Positive</u> (e.g., present)<br><u>Negative</u> (e.g., not identified, negative) |
| **Lymphatic Invasion** | <u>Positive</u> (e.g., present)<br><u>Negative</u> (e.g., not identified, negative, absent) |
| **Lymphovascular Invasion** | <u>Positive</u> (e.g., present)<br><u>Negative</u> (e.g., not identified, negative, absent) |
| **Extrathyroidal Extension** | <u>Macroscopic:</u> Evidence of macroscopic invasion<br><u>Microscopic moderate/severe:</u> Subcutaneous tissue, larynx, trachea, esophagus, nerves.<br><u>Microscopic minimal:</u> Perithyroidal adipose tissue, strap muscles (e.g., sternohyoid, omohyoid, thyrohyoid and sternohyoid)<br><u>Negative</u> (e.g., not identified) |
| **Number of Lymph Nodes Examined** | Any value in number or letters |
| **Number of Lymph Nodes Involved** | Any value in number or letters |
| **Extranodal Extension** | <u>Positive</u> (e.g., present)<br><u>Negative</u> (e.g., not identified, absent) |
| **Size of Largest Metastatic Deposit** | Any value |
| **Primary tumor TNM** | T0 (e.g., no evidence of primary tumor)<br>T1 (tumor ≤ 2 cm limited to the thyroid)<br>T2 (tumor > 2 cm but ≤4 cm limited to the thyroid)<br>T3 (tumor > 4 cm limited to the thyroid or gross ETE invading only strap muscles)<br>T4 (gross extrathyroidal extension into major neck structures) |
| **Lymph nodes TNM** | NX (lymph nodes cannot be assessed, not applicable)<br>N0 (no evidence of regional lymph nodes)<br>N1 (metastasis to regional lymph nodes) |
| **Distant Metastasis** | MX (e.g., Distant metastasis cannot be assessed)<br>M0 (no evidence of distant metastasis)<br>M1 (distant metastasis) |
| **Pathologic staging** | AJCC 8TH edition<br>AJCC 7TH edition |

| Supplemental Table 2. Extraction of numerical categories from surgical pathology reports | |
|---|---|
| **Name of Numerical Categories** | Key Strings Applied to the Unstructured Reports |
| **TumorSize** | 'tumor size: […] histologic type'<br>'forming […] the'<br>'measure […] the'<br>'in greatest dimension' |
| **SizeOfLargestMetastaticDeposit** | 'largest metastatic focus in the lymph node'<br>'lymph node metastasis:'<br>'largest metastatic focus in the lymph node:'<br>'largest metastasis measures'<br>'lymph node measures' |
| **NumberOfLymphNodesInvolved** | 'number involved:'<br>'number involved (total)'<br>'number of lymph nodes involved:' |
| **NumberOfLymphNodesExamined** | 'number examined:'<br>'number examined (total)'<br>'number of lymph nodes examined:' |

| Supplemental table 3. Detailed performance of the NLP pipeline for extraction tasks of the categories in the surgical pathology reports, using lenient method | | | | | | | | |
|---|---|---|---|---|---|---|---|---|
| Categories in Pathology Reports | Structured Pathology Reports | | | | Unstructured Pathology Reports | | | |
| | Accuracy | Precision | Recall | F1-score | Accuracy | Precision | Recall | F1-score |
| Procedure | 0.84 | 0.93 | 0.9 | 0.92 | 0.71 | 0.88 | 0.78 | 0.83 |
| Tumor Focality | 0.98 | 1.0 | 0.98 | 0.99 | 0.94 | 0.98 | 0.96 | 0.97 |
| Tumor Site | 0.89 | 0.95 | 0.94 | 0.94 | 0.83 | 0.93 | 0.89 | 0.91 |
| Tumor Size | 1.0 | 1.0 | 1.0 | 1.0 | 1.0 | 1.0 | 1.0 | 1.0 |
| Histologic Subtype | 0.71 | 0.84 | 0.82 | 0.83 | 0.74 | 0.88 | 0.82 | 0.85 |
| Margins | 0.96 | 1.0 | 0.96 | 0.98 | 0.62 | 0.78 | 0.75 | 0.76 |
| Angioinvasion (Vascular Invasion) | 0.96 | 0.99 | 0.97 | 0.98 | 1.0 | 1.0 | 1.0 | 1.0 |
| Lymphatic Invasion | 0.95 | 0.98 | 0.97 | 0.97 | 0.92 | 0.96 | 0.96 | 0.96 |
| Lymphovascular Invasion | (n.a.) * | | | | 0.94 | 1.0 | 0.94 | 0.97 |
| Extrathyroidal Extension | 0.90 | 0.96 | 0.94 | 0.95 | 0.90 | 0.96 | 0.94 | 0.95 |
| Number of Lymph Nodes Involved | 0.79 | 0.98 | 0.8 | 0.88 | 0.68 | 1.0 | 0.68 | 0.81 |
| Number of Lymph Nodes Examined | 0.79 | 0.98 | 0.8 | 0.88 | 0.90 | 1.0 | 0.90 | 0.95 |
| Size of Largest Metastatic Deposit | 0.81 | 0.98 | 0.83 | 0.90 | 0.92 | 0.98 | 0.94 | 0.96 |
| Extranodal Extension | 0.95 | 0.99 | 0.96 | 0.97 | 0.80 | 0.90 | 0.88 | 0.89 |
| Pathologic Staging | 1.0 | 1.0 | 1.0 | 1.0 | 1.0 | 1.0 | 1.0 | 1.0 |
| Primary Tumor TNM | 0.98 | 1.0 | 0.98 | 0.99 | 1.0 | 1.0 | 1.0 | 1.0 |
| Lymph Nodes TNM | 0.98 | 1.0 | 0.98 | 0.99 | 1.0 | 1.0 | 1.0 | 1.0 |
| Distant Metastasis | 0.98 | 1.0 | 0.98 | 0.99 | 1.0 | 1.0 | 1.0 | 1.0 |
| Overall | 0.91 | 0.98 | 0.93 | **0.95** | 0.88 | 0.96 | 0.91 | **0.93** |

* Lymphovascular invasion is only present in unstructured reports, thus not applicable (n.a.) in structured cohort.

| Supplemental table 4. Classification tasks assessing human-extracted features versus ground truth. | | | | | |
|---|---|---|---|---|---|
| NLP pipeline / Ground truth | High risk | Intermediate risk | Low risk | Very low risk | Total |
| High risk | **35** | 0 | 0 | 0 | 35 |
| Intermediate risk | 0 | **58** | 0 | 0 | 58 |
| Low risk | 0 | 0 | **146** | 0 | 146 |
| Very low risk | 0 | 0 | 0 | **31** | 31 |
| Total | 35 | 58 | 146 | 31 | **270** |

| Supplemental table 5. Distribution of recurrence risk stratification attributes per category. | | |
|---|---|---|
| **Category** | Attributes | No. (%) |
| **High risk** | Extranodal extension | 21 (60%) |
| | Size of largest metastatic deposit > 3cm | 12 (34.2%) |
| | Distant metastasis | 1 (2.9%) |
| | Extrathyroidal extension macroscopic | 1 (2.9%) |
| | Total | 35 |
| **Intermediate risk** | Number of lymph nodes involved > 5 | 26 (44.8%) |
| | Tumor size > 4 cm | 17 (29.3%) |
| | Angioinvasion (vascular invasion) | 7 (12.1%) |
| | Size of largest metastatic deposit between 3 cm and 1 cm | 4 (6.9%) |
| | Aggressive histology | 4 (6.9%) |
| | Total | 58 |
| **Low risk** | Tumor size between 4 cm and 1 cm | 116 (79.5%) |
| | Number of lymph nodes involved < 5 | 27 (18.5%) |
| | Extrathyroidal extension microscopic minimal | 2 (1.4%) |
| | Size of largest metastatic deposit < 1 cm | 1 (0.6%) |
| | Total | 146 |
| **Very low risk** | Tumor size < 1 cm | 31 (100%) |
| | Total | 31 |

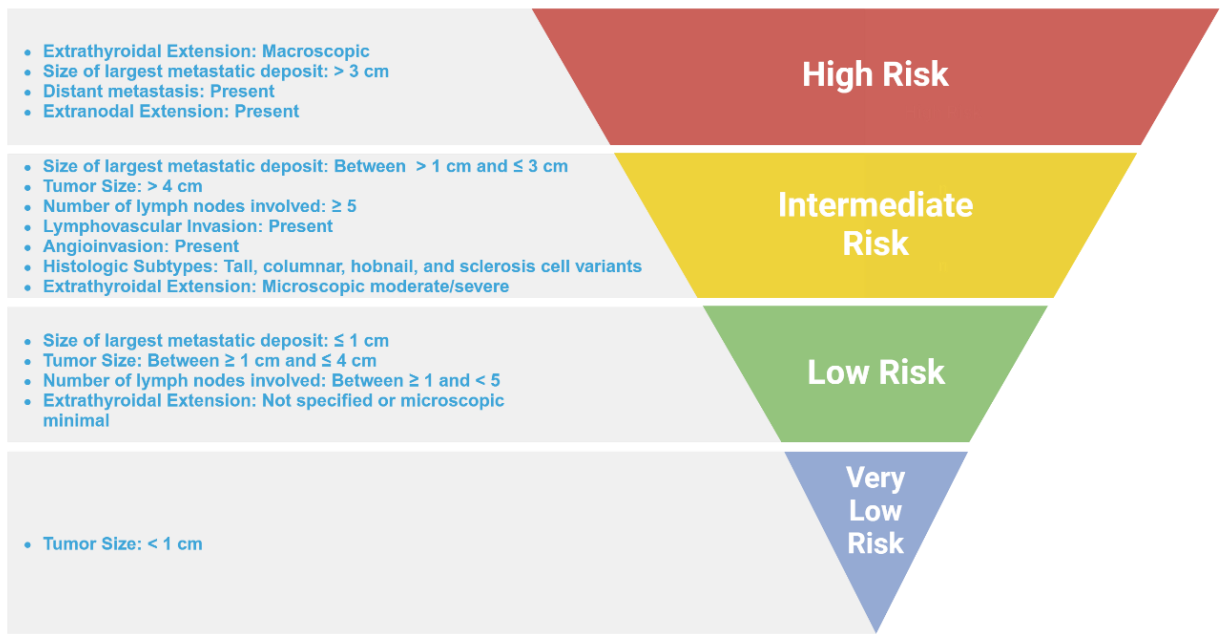

- Extrathyroidal Extension: Macroscopic
- Size of largest metastatic deposit: > 3 cm
- Distant metastasis: Present
- Extranodal Extension: Present

**High Risk**

- Size of largest metastatic deposit: Between > 1 cm and ≤ 3 cm
- Tumor Size: > 4 cm
- Number of lymph nodes involved: ≥ 5
- Lymphovascular Invasion: Present
- Angioinvasion: Present
- Histologic Subtypes: Tall, columnar, hobnail, and sclerosis cell variants
- Extrathyroidal Extension: Microscopic moderate/severe

**Intermediate Risk**

- Size of largest metastatic deposit: ≤ 1 cm
- Tumor Size: Between ≥ 1 cm and ≤ 4 cm
- Number of lymph nodes involved: Between ≥ 1 and < 5
- Extrathyroidal Extension: Not specified or microscopic minimal

**Low Risk**

- Tumor Size: < 1 cm

**Very Low Risk**

**Supplemental Figure 1: Annotation flowchart for categorizing the risk of recurrence of papillary thyroid cancer.**

SYNOPTIC REPORT
Procedure: Total thyroidectomy.
Tumor Focality: Unifocal.
Tumor Site: Right lobe.
Tumor Size: Greatest dimension: 2.8 cm.
Histologic Type: Papillary carcinoma, classic.
Margins: Involved by carcinoma.
Angioinvasion (Vascular Invasion): Not identified.
Lymphatic Invasion: Present.
Extrathyroidal Extension: Not identified.
Regional Lymph Nodes
    Number of Lymph Nodes Involved: 23.
        Level VI, levels IIA, IIB, III and IV.
    Number of Lymph Nodes Examined: 44.
        Level VI, levels II-IV, paraesophageal.
    Size of Largest Metastatic Deposit: 5.5 cm.
    Extranodal Extension: Present.
Pathologic Staging (AJCC, 8th edition)
    TNM Descriptors: Not applicable.
    Primary Tumor: pT2.
    Regional lymph nodes: pN1b.
    Distant Metastasis: Not applicable.

DIAGNOSIS:
A. Thyroid and isthmus, right thyroidectomy and isthmusectomy: Grade 1 (of 4) papillary thyroid carcinoma, follicular subtype forming two nodules (0.7 x 0.5 x 0.4 cm and 0.6 x 0.5 x 0.4 cm). Tumor is confined to the thyroid. All surgical resection margins are negative for tumor. A single (1 of 3) perithyroidal lymph nodes is positive for metastatic papillary thyroid carcinoma. Extranodal extension is not identified.

B. Thyroid, left thyroidectomy: Grade 1 (of 4) papillary thyroid carcinoma, follicular subtype, forming a microscopic focus measuring 0.1 x 0.1 x 0.1 cm located in the mid portion of the lobe. The tumor is confined to the thyroid. All surgical resection margins are negative for tumor.

C. Lymph node, bilateral neck, level VI, biopsy: Multiple (4) lymph nodes are negative for tumor.

With available surgical material [AJCC pT1N1] (7th edition, 2010).

**Supplemental Figure 2. 2a (left) Showcases structured report employing a header-based topic segmentation strategy and Figure 2b (right) illustrates an unstructured report with a topic segmentation strategy.**